\documentclass[pdflatex,sn-mathphys-num]{sn-jnl}

\usepackage{graphicx}
\usepackage{booktabs}
\usepackage{array}
\usepackage{multirow}
\usepackage{rotating} 
\usepackage{amsmath,amssymb,amsfonts}%
\usepackage{amsthm}%
\usepackage{mathrsfs}%
\usepackage[title]{appendix}%
\usepackage{xcolor}%
\usepackage{textcomp}%
\usepackage{manyfoot}%
\usepackage{booktabs}%
\usepackage{algorithm}%
\usepackage{algorithmicx}%
\usepackage{algpseudocode}%
\usepackage{listings}%
\usepackage{lmodern}
\usepackage{mathpazo}
\usepackage{xcolor}
\usepackage{soul}

\begin{document}

\title[Article Title]{HONeYBEE: Enabling Scalable Multimodal AI in Oncology Through Foundation Model–Driven Embeddings}

\author*[1,3]{\fnm{Aakash} \sur{Tripathi}}\email{aakash.tripathi@moffitt.org}
\equalcont{These authors contributed equally to this work.}
\author[1,2]{\fnm{Asim} \sur{Waqas}}\email{asim.waqas@moffitt.org}
\equalcont{These authors contributed equally to this work.}
\author[2]{\fnm{Matthew} \spfx{B.} \sur{Schabath}}\email{matthew.schabath@moffitt.org}
\author[3]{\fnm{Yasin} \sur{Yilmaz}}\email{yasiny@usf.edu}
\author[1,3]{\fnm{Ghulam} \sur{Rasool}}\email{ghulam.rasool@moffitt.org}
\affil[1]{\orgdiv{Department of Machine Learning}, \orgname{Moffitt Cancer Center \& Research Institute}, \orgaddress{\street{12902 USF Magnolia Drive}, \city{Tampa}, \postcode{33620}, \state{Florida}, \country{USA}}}
\affil[2]{\orgdiv{Departments of Cancer Epidemiology}, \orgname{Moffitt Cancer Center \& Research Institute}, \orgaddress{\street{12902 USF Magnolia Drive}, \city{Tampa}, \postcode{33620}, \state{Florida}, \country{USA}}}
\affil[3]{\orgdiv{Department of Electrical Engineering}, \orgname{University of South Florida}, \orgaddress{\street{4202 E Fowler Ave}, \city{Tampa}, \postcode{33620}, \state{Florida}, \country{USA}}}

\abstract{HONeYBEE (Harmonized ONcologY Biomedical Embedding Encoder) is an open-source framework that integrates multimodal biomedical data for oncology applications. It processes clinical data (structured and unstructured), whole-slide images, radiology scans, and molecular profiles to generate unified patient-level embeddings using domain-specific foundation models and fusion strategies. These embeddings enable survival prediction, cancer-type classification, patient similarity retrieval, and cohort clustering. Evaluated on 11,400+ patients across 33 cancer types from The Cancer Genome Atlas (TCGA), clinical embeddings showed the strongest single-modality performance with 98.5\% classification accuracy and 96.4\% precision@10 in patient retrieval. They also achieved the highest survival prediction concordance indices across most cancer types. Multimodal fusion provided complementary benefits for specific cancers, improving overall survival prediction beyond clinical features alone. Comparative evaluation of four large language models revealed that general-purpose models like Qwen3 outperformed specialized medical models for clinical text representation, though task-specific fine-tuning improved performance on heterogeneous data such as pathology reports.}


\maketitle

\section{Introduction}\label{introduction}

Recent advances in computational oncology have been fueled by the increasing digitization of diverse biomedical data, including structured clinical variables (such as demographics, tumor staging, and laboratory results), unstructured clinical narratives (such as pathology reports, radiology reports, and physician notes), medical imaging (radiology scans and whole-slide images or WSI), and high-dimensional molecular profiles. \cite{boehm2022harnessing, paverd2024radiology, Vanguri2022, waqas2024multimodal, waqas2024digital, Boehm2022}. This wealth of multimodal data offers unprecedented opportunities to improve patient stratification, predict treatment response, and model disease progression \cite{Boehm2022, Vanguri2022, paverd2024radiology}. In parallel, the adaptation of deep learning techniques from computer vision and natural language processing has enabled powerful solutions in these domains \cite{lipkova2022artificial, Vanguri2022}. However, a fundamental challenge remains: the absence of robust, generalizable methods for integrating these heterogeneous data sources into unified representations that capture the biological complexity of cancer and support predictive modeling \cite{MINDS}.

Although large-scale biomedical data is increasingly available and actively analyzed in oncology, it remains fragmented across distinct modalities, such as clinical data (structured variables and unstructured narratives), radiological and pathological imaging, and molecular profiles, which are typically processed separately. This siloed approach limits the ability to integrate complementary information across modalities for unified, patient-centered analysis \cite{TCGA, MINDS}. Availability of large-scale datasets and advances in self-supervised learning have enabled the development of foundation models (FMs) \cite{foundationmodels, waqas2023revolutionizing, boehm2022harnessing}. These models, pretrained on text, imaging, or molecular data, have advanced feature extraction within individual modalities by learning latent representations that capture domain-specific patterns. These modality-specific embeddings can be adapted for downstream oncology tasks such as cancer classification or overall survival (OS) prediction. However, in practice, these models are typically applied within single- or dual-modality workflows, leaving the complementary information across modalities underutilized \cite{Hartsock2024}. While multimodal data availability continues to expand in oncology, a critical bottleneck remains: the absence of standardized, scalable frameworks that integrate modality-specific embeddings into unified, patient-level representations that capture multimodal patient similarity and support downstream oncology tasks.

We hypothesize that integrating FM-derived embeddings from multiple data modalities can yield richer and more clinically informative patient representations, particularly in settings where clinical data are incomplete or less structured. Rather than relying solely on model scaling or increasing parameter counts, we propose that fusing complementary information from diverse biomedical data types offers a powerful, orthogonal approach to enhance predictive performance in oncology. To test this hypothesis, we present \textbf{HONeYBEE} or \textbf{H}armonized \textbf{ON}colog\textbf{Y} \textbf{B}iomedical \textbf{E}mbedding \textbf{E}ncoder (\url{https://lab-rasool.github.io/HoneyBee/}). HONeYBEE is an open-source framework that generates individual patient-level embeddings from (i) structured and unstructured clinical data, (ii) pathology reports, (iii) radiologic images, (iv) WSIs, and (v) molecular profiles using modality-specific FMs. HONeYBEE integrates these embeddings via concatenation, mean pooling, and Kronecker product fusion strategies to create unified, multimodal representations optimized for downstream oncology tasks, including cancer subtype classification, patient clustering, OS prediction, and patient similarity retrieval.

While numerous models and pipelines exist for analyzing clinical, imaging, and molecular data, most current tools remain modality-specific and lack the flexibility to support unified, end-to-end multimodal workflows \cite{steyaert2023multimodal, boehm2022harnessing}. Existing methods are typically implemented as isolated codebases with rigid dependencies, domain-specific interfaces, and limited extensibility, which complicates reproducibility and impedes multimodal experimentation \cite{nicora2020integrated}. Moreover, the absence of standardized pipelines for modality-specific embedding generation, harmonization, and flexible fusion introduces substantial technical barriers, slowing the development of clinically meaningful AI models \cite{cooper2023machine}. Addressing these limitations requires not only access to multimodal data but also modular infrastructure capable of generating, integrating, and utilizing diverse patient-level embeddings in scalable, reproducible ways.

\begin{figure}[htpb]
    \centering
    \includegraphics[width=\textwidth]{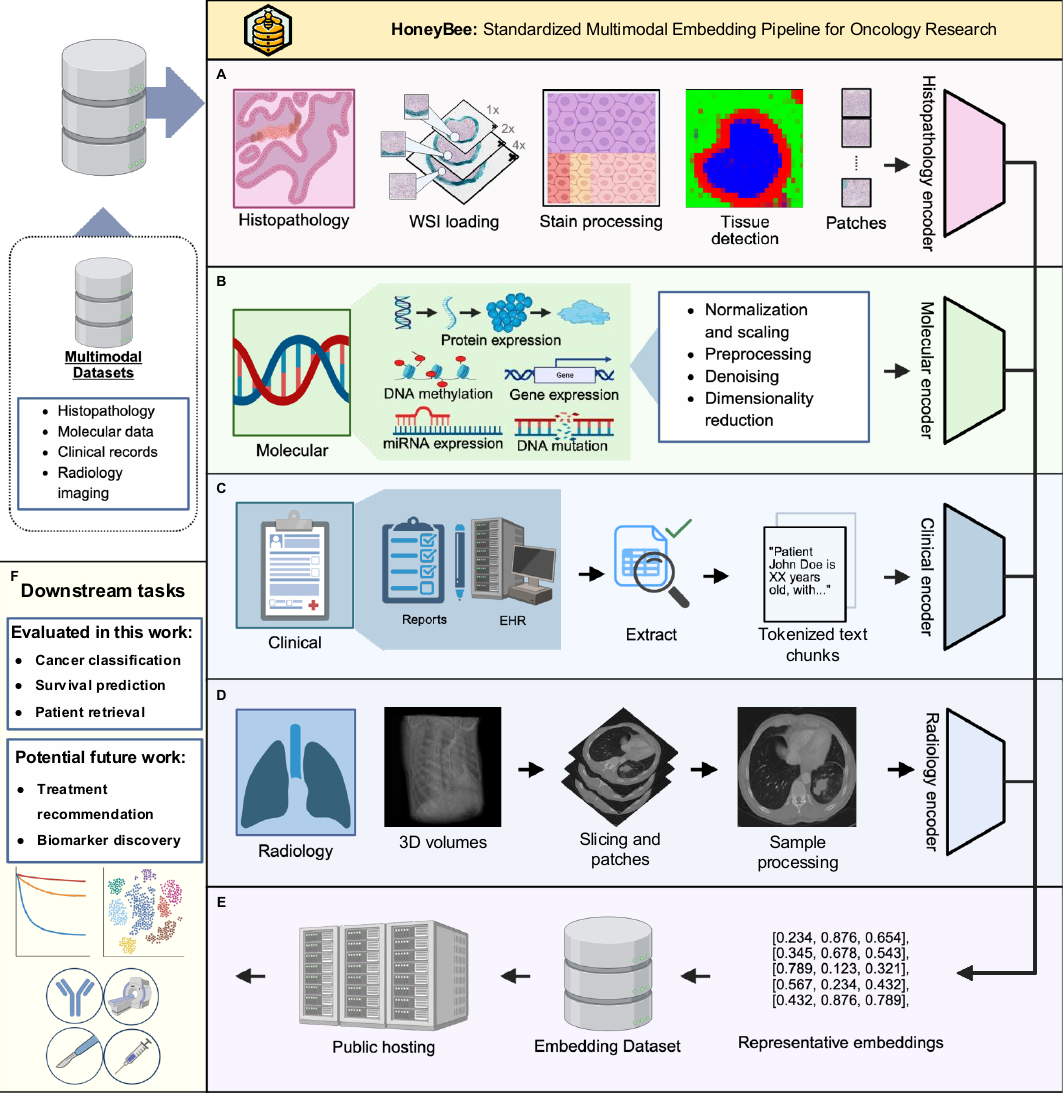}
    \caption{\textbf{Overview of the HONeYBEE framework.} HONeYBEE provides a standardized, scalable framework for generating and integrating multimodal embeddings from oncology datasets. Multimodal datasets from public repositories, such as the NCI CRDC \cite{CRDC}, including the PDC \cite{PDC}, GDC \cite{GDC}, IDC \cite{IDC}, and TCIA \cite{TCIA}, as well as private institutional datasets, can be processed through dedicated modality-specific pipelines. \textbf{(A)} Histopathology images are processed via WSI loading, stain normalization, tissue detection, and patch extraction prior to the extraction of embeddings. \textbf{(B)} Molecular data undergo cleaning and normalization across multiple omics layers, including DNA methylation, gene expression, and somatic mutations. \textbf{(C)} Clinical text, including pathology and radiology reports and structured data from the electronic health record (EHR), is extracted and tokenized into coherent text segments for downstream processing. \textbf{(D)} Radiology images are processed using volumetric slicing, patch extraction, and modality-specific preprocessing optimized for downstream tasks. \textbf{(E)} Embeddings, fixed-length feature vectors that capture the most informative patterns from each data modality, are generated using FMs. HONeYBEE supports both pretrained, publicly available, and custom FMs via the Hugging Face ecosystem. All extracted embeddings and processed datasets are publicly released to promote reproducibility, sharing, and external validation. \textbf{(F)} The multimodal embeddings serve as a standardized input for downstream oncology tasks, such as those evaluated in this work (cancer classification, OS prediction, patient retrieval) and potential future directions (such as treatment recommendation and biomarker discovery).}
    \label{fig:overview}
\end{figure}

HONeYBEE directly addresses this gap by providing a modular, open-source framework for multimodal embedding generation and integration. Built around domain-specific FMs, HONeYBEE supports the standardized preprocessing and representation of five key oncology data modalities \cite{UNI, gatortron, REMEDIS, SeNMo, radimagenet, MINDS}. Each modality is processed through dedicated pipelines, producing modality-specific embeddings optimized for downstream use. These embeddings can then be integrated using flexible fusion strategies. Through a simplified API and modular design, HONeYBEE enables oncology researchers to easily incorporate new FMs, compare embedding strategies, and deploy multimodal embeddings for new tasks. By abstracting away complexities related to data harmonization, preprocessing, and embedding fusion, HONeYBEE facilitates scalable, reproducible, and clinically meaningful multimodal analysis. An overview of the framework's architecture and supported modalities is shown in Fig. \ref{fig:overview}.

To further support adoption and interoperability, HONeYBEE is designed for seamless integration with widely used biomedical data repositories and machine learning platforms. It supports direct data ingestion from resources such as the NCI Cancer Research Data Commons (CRDC), Proteomics Data Commons (PDC), Genomic Data Commons (GDC), Imaging Data Commons (IDC), and The Cancer Imaging Archive (TCIA). The framework is fully compatible with PyTorch, Hugging Face, and FAISS, and includes pretrained FMs along with pipelines for adding new models and modalities. Unlike existing single-modality tools or highly customized pipelines, HONeYBEE offers flexible deployment, standardized embedding workflows, and minimal-code implementation of state-of-the-art techniques \cite{soenksen2022integrated, pahud2024orchestrating, chang2025continuous}. In the sections that follow, we evaluate HONeYBEE’s performance across key oncology tasks, including clustering, cancer subtype classification, OS prediction, and patient similarity retrieval, using the TCGA dataset to demonstrate its effectiveness as a generalizable platform for multimodal cancer research.

\section{Results}\label{results}

We evaluated HONeYBEE using multimodal patient-level data from The Cancer Genome Atlas (TCGA), encompassing 11,428 patients across 33 cancer types. Available data modalities included clinical text (11,428 patients), molecular profiles (13,804 samples from 10,938 patients), pathology reports (11,108 patients), WSIs (8,060 patients), and radiologic images (1,149 patients). This heterogeneous, incomplete modality availability reflects real-world clinical data constraints and allowed evaluation of HONeYBEE’s robustness in handling missing data.

HONeYBEE-generated embeddings were assessed across four core downstream tasks: cancer type classification, patient similarity retrieval, cancer-type clustering, and OS prediction. Analyses evaluated both modality-specific embeddings and integrated multimodal embeddings generated using fusion strategies such as concatenation, mean pooling, and Kronecker product. The modular design of HONeYBEE accommodated patients with missing modalities without requiring complete-case cohorts.

\subsection{Modality-Specific Embedding Generation}

HONeYBEE supports generating modality-specific embeddings from five primary data types using state-of-the-art FMs. This section describes the default models integrated into HONeYBEE for each data modality, explaining the rationale behind their selection and their specific capabilities. While we present and evaluate specific models in our analysis, any FMs from Hugging Face can be added in HONeYBEE.

For clinical text and pathology reports, HONeYBEE supports multiple language models: GatorTron \cite{gatortron}, Qwen3 \cite{qwen}, Med-Gemma \cite{medgemma}, and Llama-3.2 \cite{llama}. For the primary analyses presented in this paper (cancer type classification, survival prediction, and patient retrieval), we used GatorTron embeddings due to its specialized training on clinical text. The comparative evaluation of all four language models (GatorTron, Qwen3, Med-Gemma, and Llama-3.2) is presented separately in Section \ref{subsec:comparative_llm_eval} to demonstrate the framework's flexibility in supporting diverse architectural paradigms.

For WSIs embedding generation, HONeYBEE supports UNI (ViT-L/16, 307M parameters), UNI2-h (ViT-g/14, 632M parameters), and Virchow2 (DINOv2, 1.1B parameters) \cite{UNI, zimmermann2024virchow2}. UNI provides 1,024-dimensional embeddings optimized for efficiency, making it suitable for large-scale processing with limited computational resources. UNI2-h, with its larger architecture, offers enhanced feature extraction capabilities, while Virchow2, built on the DINOv2 framework, provides the most comprehensive representations through self-supervised learning on diverse pathology datasets. For the primary analyses in this paper, we used UNI embeddings, while the comparative evaluation of staining impact across all three models is presented in Supplementary Materials (see Supplementary Note - Staining impact of performance, Supplementary Figures 10 and 11, and Supplementary Tables 7 and 8).

For the radiological imaging data, HONeYBEE supports RadImageNet, a convolutional neural network (CNN) pre-trained on over four million medical images across CT, MRI, and PET modalities \cite{radimagenet}. The model's multi-modality pretraining enables it to handle the diverse imaging protocols commonly encountered in oncology, from contrast-enhanced CT scans to functional PET imaging. RadImageNet embeddings were used for all radiology analyses presented in this paper.

\begin{table}[htbp]
  \centering
  \caption{Performance of Multimodal Fusion Strategies vs. Single-Modality Embeddings for Cancer-Type Clustering}
  \label{tab:nmi_ami}
    \begin{tabular}{@{}lcc@{}}
      \toprule
      \textbf{Method} & \textbf{Normalized Mutual Information} & \textbf{Adjusted Mutual Information} \\
      & \textbf{(NMI)} & \textbf{(AMI)} \\
      \midrule
      \multicolumn{3}{l}{\textit{Individual Modalities}} \\
      Clinical (best single modality) & 0.7448 & 0.702 \\
      Molecular & 0.3543 & 0.249 \\
      Radiology & 0.2719 & 0.228 \\
      Pathology Reports & 0.4352 & 0.341 \\
      Whole Slide Images (WSI) & 0.1322 & 0.056 \\
      \midrule
      \multicolumn{3}{l}{\textit{Multimodal Fusion}} \\
      Concatenation & 0.4440 & 0.347 \\
      Mean Pooling & 0.4350 & 0.336 \\
      Kronecker Product & 0.4053 & 0.320 \\
      \bottomrule
    \end{tabular}
\end{table}

For the molecular data, HONeYBEE includes SeNMo \cite{SeNMo}. SeNMo is a self-normalizing deep learning encoder specifically designed for high-dimensional multi-omics data, including gene expression, DNA methylation, somatic mutations, miRNA, and protein expression \cite{SeNMo}. The self-normalizing properties of SeNMo ensure stable training even with the diverse scales and distributions present in multi-omics data. SeNMo embeddings were used for all molecular analyses presented in this paper.

Modality-specific embeddings are generated using standardized pipelines provided in the HONeYBEE framework, which is fully compatible with the Hugging Face ecosystem. This design choice enables easy integration of new FMs as they become available. HONeYBEE's API abstracts the complexity of model-specific preprocessing requirements while maintaining flexibility. The resulting patient-level feature vectors, along with associated metadata, have been publicly released via Hugging Face repositories including TCGA (\url{https://huggingface.co/datasets/Lab-Rasool/TCGA}), CGCI (\url{https://huggingface.co/datasets/Lab-Rasool/CGCI}), Foundation Medicine (\url{https://huggingface.co/datasets/Lab-Rasool/FM}), CPTAC (\url{https://huggingface.co/datasets/Lab-Rasool/CPTAC}), and TARGET (\url{https://huggingface.co/datasets/Lab-Rasool/TARGET}).

\subsection{Multimodal Embedding Integration and Analysis}
For multimodal integration, we implemented three fusion strategies within HONeYBEE to accommodate heterogeneous data availability across 11,424 patients (99.97\% of the cohort) who had at least two available modalities: (1) concatenation, which joins available embeddings while preserving modality-specific information; (2) mean pooling, which averages embeddings after padding to a common dimension; and (3) Kronecker product, which captures pairwise interactions between modalities. To evaluate the resulting multimodal embedding spaces, we quantified cancer-type separability using normalized mutual information (NMI) and adjusted mutual information (AMI), which respectively measure the alignment of embedding clusters with known cancer-type labels and clustering quality adjusted for chance.

Table~\ref{tab:nmi_ami} presents both clustering metrics (NMI and AMI) for single and multimodal embeddings and for all three integration strategies. Surprisingly, clinical embeddings alone yielded the strongest cancer-type clustering, with an NMI of 0.7448 and AMI of 0.702, outperforming both other single modalities and all multimodal fusion strategies. This result likely reflects the curated nature of clinical documentation in TCGA, where key diagnostic variables, such as cancer subtype, stage, grade, and molecular markers, are extracted and recorded by clinical experts, effectively summarizing information that may be dispersed across raw radiology, pathology, and molecular data. However, all three multimodal fusion approaches, concatenation, mean pooling, and Kronecker product, outperformed weaker single modalities such as molecular, radiology, and WSI embeddings. Among fusion methods, concatenation achieved the best clustering performance, with an NMI of 0.4440 and AMI of 0.347.

Although multimodal fusion did not surpass clinical embeddings in clustering performance, it provided a robust approach to integrating heterogeneous data types and accommodating missing modalities. Visualization of the concatenated multimodal embedding space demonstrated clearer cancer-type separation compared to weaker single modalities, such as radiology and WSI (Supplementary Figures 6-8). While clinical text remained the dominant signal for cancer-type differentiation, multimodal fusion effectively combined information from lower-performing modalities, supporting more comprehensive patient-level representations in cases with limited clinical data.

We further evaluated the clustering behavior of each individual modality to establish a baseline for cancer-type separability. Clinical embeddings demonstrated the strongest performance, with clearly defined boundaries between 33 distinct cancer types and achieving 90.21\% classification accuracy using a simple random forest classifier. Molecular embeddings (13,804 samples) showed structured but overlapping clusters, reflecting biological similarities among related cancers. Pathology report embeddings (11,108 samples) exhibited moderate cancer-type clustering, while radiology embeddings (1,149 samples) displayed diffuse clusters, consistent with limited data availability and lower cancer-type specificity. Whole-slide image embeddings (8,060 samples) showed minimal clustering, likely due to slide-level variability and indirect representation of cancer-type features. Detailed t-SNE visualizations for all modalities are provided in the Supplementary Materials (Supplementary Figures 1-6 and Supplementary Note - Embeddings visualization).

\subsection{Evaluation of Multimodal Embeddings on Downstream Tasks}

We evaluated HONeYBEE-generated embeddings on three key downstream tasks: OS prediction, cancer type classification, and patient similarity retrieval. These tasks were selected to assess the generalizability and practical utility of both modality-specific and multimodal embeddings. Given the heterogeneous availability of clinical, radiologic, WSI, and molecular data across patients, we evaluated both standalone and integrated embeddings to assess robustness under real-world data constraints.


\begin{table}[htbp]
    \centering
    \caption{Comparison of survival prediction methods on TCGA-BRCA and TCGA-BLCA datasets. Results are reported as concordance index (mean $\pm$ std).}
    \label{tab:survival_results}
    \begin{tabular}{@{}llcc@{}}
        \toprule
        \textbf{Category} & \textbf{Method} & \textbf{BRCA} & \textbf{BLCA} \\
        \midrule
        \multirow{6}{*}{\textbf{Baseline Methods}} 
        & SurvPath \cite{survpath} & 0.655 ± 0.089 & 0.625 ± 0.056 \\
        & ABMIL (KP) \cite{abmil} & 0.615 ± 0.083 & 0.566 ± 0.038 \\
        & MCAT \cite{mcat} & 0.652 ± 0.117 & 0.598 ± 0.094 \\
        & MOTCat \cite{motcat} & 0.600 ± 0.095 & 0.596 ± 0.079 \\
        & Porpoise \cite{porpoise} & 0.652 ± 0.042 & 0.636 ± 0.024 \\
        & PathOmic \cite{pathomic} & --- & 0.586 ± 0.062 \\
        \midrule
        \multicolumn{4}{l}{\textit{Individual Modalities}} \\
        \midrule
        \multirow{3}{*}{\textbf{Clinical Features}} 
        & Cox & 0.920 ± 0.017 & 0.823 ± 0.011 \\
        & RSF & 0.845 ± 0.020 & 0.799 ± 0.026 \\
        & DeepSurv & 0.928 ± 0.013 & 0.842 ± 0.018 \\
        \midrule
        \multirow{3}{*}{\textbf{Pathology Report Features}} 
        & Cox & 0.495 ± 0.056 & 0.556 ± 0.044 \\
        & RSF & 0.511 ± 0.066 & 0.534 ± 0.038 \\
        & DeepSurv & 0.497 ± 0.047 & 0.561 ± 0.025 \\
        \midrule
        \multirow{3}{*}{\textbf{Molecular Features}} 
        & Cox & 0.525 ± 0.066 & 0.501 ± 0.026 \\
        & RSF & 0.426 ± 0.061 & 0.536 ± 0.045 \\
        & DeepSurv & 0.520 ± 0.012 & 0.468 ± 0.015 \\
        \midrule
        \multirow{3}{*}{\textbf{WSI Features}} 
        & Cox & 0.461 ± 0.051 & 0.519 ± 0.053 \\
        & RSF & 0.491 ± 0.049 & 0.524 ± 0.036 \\
        & DeepSurv & 0.464 ± 0.051 & 0.517 ± 0.048 \\
        \midrule
        \multicolumn{4}{l}{\textit{Multimodal Fusion}} \\
        \midrule
        \multirow{3}{*}{\textbf{Concatenation}} 
        & Cox & 0.767 ± 0.052 & 0.799 ± 0.022 \\
        & RSF & 0.808 ± 0.029 & 0.790 ± 0.028 \\
        & DeepSurv & 0.847 ± 0.040 & 0.820 ± 0.015 \\
        \midrule
        \multirow{3}{*}{\textbf{Mean Pooling}} 
        & Cox & 0.589 ± 0.042 & 0.648 ± 0.041 \\
        & RSF & 0.592 ± 0.083 & 0.577 ± 0.061 \\
        & DeepSurv & 0.666 ± 0.048 & 0.689 ± 0.019 \\
        \midrule
        \multirow{3}{*}{\textbf{Kronecker Product}} 
        & Cox & 0.728 ± 0.030 & 0.709 ± 0.055 \\
        & RSF & 0.625 ± 0.068 & 0.705 ± 0.028 \\
        & DeepSurv & 0.758 ± 0.023 & 0.721 ± 0.018 \\
        \bottomrule
    \end{tabular}
\end{table}

We performed survival analysis stratified across all 33 TCGA cancer types, training models on each cancer type individually. For each cancer type and modality combination, we performed 5-fold cross-validation with stratification based on survival outcomes and censoring status to ensure representative distributions across folds. This stratification approach maintains balanced representation of both event rates and follow-up durations in each fold, providing reliable concordance index estimates. For cancer types with limited samples, we merged similar cancers (e.g., TCGA-COAD and TCGA-READ for colorectal cancers) to ensure model training with adequate sample sizes. For each cancer type, we trained three widely used survival models: Cox proportional hazards (CoxPH), Random Survival Forests (RSF), and DeepSurv \cite{deepsurv}. Each model was trained using both single-modality embeddings (clinical, pathology reports, molecular, WSI, and radiology images, where available) and three types of multimodal embeddings: concatenated, mean-pooled, and Kronecker product features. Predictive performance was measured using the concordance index (C-index), which quantifies the agreement between predicted risk scores and observed survival times.

Table \ref{tab:survival_results} presents a comparison of HONeYBEE embeddings against established baseline methods for survival prediction in two representative cancer types: breast cancer (TCGA-BRCA) and bladder cancer (TCGA-BLCA). Baseline models, including SurvPath \cite{survpath}, ABMIL \cite{abmil}, MCAT \cite{mcat}, MOTCat \cite{motcat}, Porpoise \cite{porpoise}, and PathOmic \cite{pathomic}, achieved concordance indices (C-indices) ranging from 0.566 to 0.655, representing the current state-of-the-art in multimodal survival prediction. In contrast, HONeYBEE embeddings demonstrated substantially superior performance. Clinical embeddings achieved the highest single-modality results using Cox models, outperforming all baseline methods and reflecting the rich prognostic information captured from structured clinical data. This likely reflects the curated nature of TCGA clinical documentation, where key diagnostic variables, such as cancer subtype, stage, grade, and molecular markers, are systematically extracted and recorded by experts, effectively summarizing prognostic signals that may remain distributed across raw imaging, pathology, and molecular data. While multimodal fusion via concatenation produced strong results, it did not surpass clinical features alone within this curated dataset. 

We consider that multimodal fusion holds critical importance in real-world oncology practice, where clinical documentation may be incomplete, inconsistently structured, or lacking detailed annotations. In such settings, integrating complementary signals from radiology, pathology, and molecular data can compensate for missing or unreliable clinical narratives. While TCGA highlights the dominant role of clinical features in expert-curated research datasets, our findings emphasize the potential of multimodal fusion to develop more robust and generalizable survival prediction models applicable to real-world, less-curated healthcare environments..

Extending the analysis to all 33 TCGA cancer types revealed considerable heterogeneity in survival prediction performance across cancers (Supplementary Tables 1-6 and Supplementary Note - Overall survival) revealed considerable heterogeneity in survival prediction performance across cancers. Clinical embeddings achieved C-indices above 0.80 for 27 of 33 cancer types, reaffirming their dominant prognostic role within TCGA. Notable examples included TCGA-PCPG (0.996 ± 0.007), TCGA-THCA (0.985 ± 0.003), and TCGA-UCEC (0.935 ± 0.012). However, certain malignancies were better characterized by other modalities; for instance, molecular embeddings achieved superior performance in TCGA-KICH (0.725 ± 0.173), suggesting a stronger molecular basis for prognosis in this cancer type. Pathology report and WSI embeddings contributed moderate predictive value across most cancers, while radiology embeddings were limited by sample availability, being present for only four cancer types.

Importantly, multimodal fusion demonstrated cancer-type-specific benefits. While concatenation generally yielded the best multimodal performance, it did not consistently outperform clinical embeddings alone. However, for specific cancers, multimodal integration enhanced prognostic accuracy beyond what clinical features captured alone. For example, in TCGA-UCS, multimodal fusion improved C-index from 0.794 ± 0.101 (clinical) to 0.836 ± 0.070; in TCGA-UVM, from 0.844 ± 0.042 to 0.860 ± 0.084; and in TCGA-THYM, from 0.978 ± 0.032 to 0.983 ± 0.033. These improvements suggest that, for certain cancer types, integrating complementary information from multiple modalities enables more accurate survival predictions than relying solely on clinical documentation.

To assess the risk stratification performance of our survival models, we examined Kaplan-Meier survival curves based on model-predicted risk groups (Supplementary Figures 7-9) illustrate the performance of Cox proportional hazards, RFS, and DeepSurv models, respectively, for bladder cancer (TCGA-BLCA). These visualizations demonstrate how clinical embeddings consistently achieve superior patient stratification compared to other modalities, with clear separation between low, medium, and high-risk groups (log-rank p $<$ 0.001 for clinical features across all models). 


HONeYBEE embeddings were evaluated for cancer type classification using Random Forest models trained on the full TCGA dataset with all five data modalities. We conducted 10 classification experiments with different random seeds, using an 80/20 train-test split with stratification to maintain class balance. Each Random Forest model comprised 100 estimators.

\begin{table}[h]
\centering
\caption{Cancer type classification performance across modalities and fusion methods. Results show mean ± standard deviation across 10 independent runs.}
\label{tab:classification_performance}
    \begin{tabular}{lcccc}
        \toprule
        \textbf{Modality/Method} & \textbf{Accuracy (\%)} & \textbf{F1-Score (\%)} & \textbf{Precision (\%)} & \textbf{Recall (\%)} \\
        \midrule
        \multicolumn{5}{l}{\textit{Individual Modalities}} \\
        Clinical & 98.54 ± 0.32 & 98.51 ± 0.32 & 98.58 ± 0.31 & 98.54 ± 0.32 \\
        Pathology Reports & 78.67 ± 0.60 & 77.13 ± 0.64 & 78.52 ± 0.80 & 78.67 ± 0.60 \\
        Molecular & 56.73 ± 0.79 & 55.73 ± 0.72 & 57.06 ± 0.81 & 56.73 ± 0.79 \\
        Radiology & 47.78 ± 3.51 & 44.11 ± 3.01 & 42.94 ± 3.19 & 47.78 ± 3.51 \\
        WSI & 28.41 ± 0.82 & 24.83 ± 0.79 & 25.33 ± 1.06 & 28.41 ± 0.82 \\
        \midrule
        \multicolumn{5}{l}{\textit{Multimodal Fusion}} \\
        Concatenation & \textbf{98.72 ± 0.21} & \textbf{98.66 ± 0.22} & \textbf{98.80 ± 0.20} & \textbf{98.72 ± 0.21} \\
        Kronecker Product & 98.50 ± 0.14 & 98.40 ± 0.19 & 98.62 ± 0.12 & 98.50 ± 0.14 \\
        Mean Pooling & 91.97 ± 0.58 & 91.30 ± 0.65 & 91.93 ± 0.63 & 91.97 ± 0.58 \\
        \bottomrule
    \end{tabular}
\end{table}

Table~\ref{tab:classification_performance} summarizes the cancer type classification performance across individual modalities and multimodal fusion strategies. Performance varied substantially across modalities, reflecting differences in the diagnostic information available within each data type. Clinical embeddings achieved the highest performance among single modalities, underscoring the rich diagnostic content captured in clinical narratives. Pathology reports also demonstrated strong discriminative power, while molecular profiles provided moderate performance. In contrast, imaging modalities, radiology and WSI, performed considerably lower, likely due to greater intra-class variability and the indirect relationship between imaging features and cancer type.

Notably, multimodal fusion methods demonstrated the superiority of integrative approaches, with concatenation achieving the highest overall accuracy, representing a marginal but consistent improvement over the best individual modality. The Kronecker product fusion achieved comparable performance, suggesting that modeling cross-modal interactions provides an additional discriminative signal. Mean pooling also showed strong performance, though lower than the other fusion methods, indicating that while some modality-specific information is lost through averaging, the method still captures substantial shared signal across modalities. The remarkably small standard deviations across runs ($<1\%$ for all methods) demonstrate the exceptional stability of our embedding representations.

We evaluated the semantic quality of HONeYBEE embeddings through similarity-based retrieval using Facebook AI Similarity Search (FAISS) open-source library, which enables efficient nearest-neighbor search across dense vector spaces \cite{faiss}. For each modality, retrieval precision\textit{@k} was calculated, measuring the proportion of patients with the same cancer type among the $k$ nearest neighbors. Higher precision\textit{@k} values indicate that embeddings effectively capture phenotypic similarity relevant to cohort matching and cancer-type discrimination.

Table \ref{tab:retrieval_results} summarizes the retrieval performance across all modalities and fusion methods. Clinical embeddings demonstrated superior retrieval capability, substantially outperforming other individual modalities. In contrast, WSI embeddings showed the poorest performance, likely due to their slide-level rather than patient-level representation. All multimodal fusion methods underperformed relative to clinical embeddings alone, with concatenation achieving the best fusion performance.

\begin{table}[htbp]
    \centering
    \caption{Retrieval performance metrics across modalities and fusion methods. Results show precision@k values, clustering and retrieval-based AMI, and failure rates for cancer-type similarity retrieval.}
    \label{tab:retrieval_results}
    \begin{tabular}{@{}lccccccc@{}}
        \toprule
        \textbf{Modality} & \textbf{Samples} & \textbf{P@5} & \textbf{P@10} & \textbf{P@20} & \textbf{Clustering} & \textbf{Retrieval} & \textbf{Failure\%} \\
        & & & & & \textbf{AMI} & \textbf{AMI@10} & \\
        \midrule
        \multicolumn{8}{l}{\textit{Individual Modalities}} \\
        Clinical & 10,857 & 0.976 & 0.964 & 0.947 & 0.702 & 0.868 & 3.6 \\
        Molecular & 13,804 & 0.392 & 0.350 & 0.310 & 0.249 & 0.232 & 65.0 \\
        Radiology & 1,149 & 0.373 & 0.343 & 0.310 & 0.228 & 0.272 & 66.1 \\
        Pathology Reports & 10,857 & 0.746 & 0.703 & 0.651 & 0.341 & 0.585 & 29.7 \\
        Whole Slide Images & 8,060 & 0.155 & 0.143 & 0.132 & 0.056 & 0.041 & 85.7 \\
        \midrule
        \multicolumn{8}{l}{\textit{Multimodal Fusion}} \\
        Concatenation & 11,341 & 0.504 & 0.461 & 0.418 & 0.347 & 0.408 & 53.9 \\
        Mean Pooling & 11,341 & 0.492 & 0.446 & 0.403 & 0.336 & 0.382 & 55.4 \\
        Kronecker Product & 11,341 & 0.284 & 0.269 & 0.253 & 0.320 & 0.320 & 73.1 \\
        \bottomrule
    \end{tabular}
\end{table}

These results suggest that integrating weaker modalities diluted the strong semantic signal present in clinical embeddings, reducing retrieval precision despite modest improvements in clustering metrics (e.g., AMI). This highlights an important distinction between classification and retrieval tasks: while multimodal integration enhances overall class separation, clinical embeddings alone provide the most semantically meaningful representations for similarity-based patient retrieval.

The failure analysis, given in table \ref{tab:retrieval_confusion}, revealed specific patterns of confusion between cancer types. For clinical embeddings, the most commonly confused pairs included READ as COAD (562 cases), LUAD as LUSC (480 cases), and LUSC as LUAD (401 cases), reflecting known biological similarities between these cancer types. WSI embeddings showed extreme confusion rates, with 85.7\% of queries failing to retrieve same-type patients in the top 10 results.

\begin{table}[htbp]
  \centering
  \caption{Top confusion patterns in retrieval failures for each modality. Values indicate the number of times a query cancer type (row) retrieved a different cancer type (column) in the top 10 results.}
  \label{tab:retrieval_confusion}
  \begin{tabular}{@{}lllcc@{}}
      \toprule
      \textbf{Modality} & \textbf{Query Type} & \textbf{Retrieved Type} & \textbf{Count} & \textbf{Biological Relevance} \\
      \midrule
      \multirow{5}{*}{Clinical} 
      & READ & COAD & 562 & Colorectal \\
      & LUAD & LUSC & 480 & Lung \\
      & LUSC & LUAD & 401 & Lung \\
      & COAD & READ & 371 & Colorectal \\ 
      & KIRP & KIRC & 292 & Kidney \\
      \midrule
      \multirow{5}{*}{Molecular}
      & GBM & LAML & 1,369 & None \\
      & LAML & GBM & 1,215 & None \\
      & UCEC & BRCA & 1,181 & None \\
      & COAD & BRCA & 965 & None \\
      & KIRC & BRCA & 952 & None \\
      \midrule
      \multirow{5}{*}{Pathology} 
      & LUAD & LUSC & 1,634 & Lung \\
      & LUSC & LUAD & 1,621 & Lung \\
      & LGG & GBM & 899 & Brain \\
      & KIRP & KIRC & 858 & Kidney \\ 
      & READ & COAD & 820 & Colorectal \\ 
      \midrule
      \multirow{5}{*}{Radiology}
      & BRCA & UCEC & 432 & None \\
      & OV & KIRC & 382 & None \\
      & KIRC & OV & 294 & None \\
      & KIRC & LIHC & 272 & None \\
      & LIHC & KIRC & 267 & None \\
      \midrule
      \multirow{5}{*}{WSI}
      & GBM & KIRC & 899 & None \\
      & BRCA & BLCA & 805 & None \\
      & BLCA & BRCA & 783 & None \\
      & LGG & BRCA & 763 & None \\
      & PRAD & BRCA & 716 & None \\
      \bottomrule
  \end{tabular}
\end{table}

\subsection{Comparative Evaluation of Language Models for Text Embeddings}\label{subsec:comparative_llm_eval}

\begin{figure}[htbp]
    \centering
    \includegraphics[width=\linewidth]{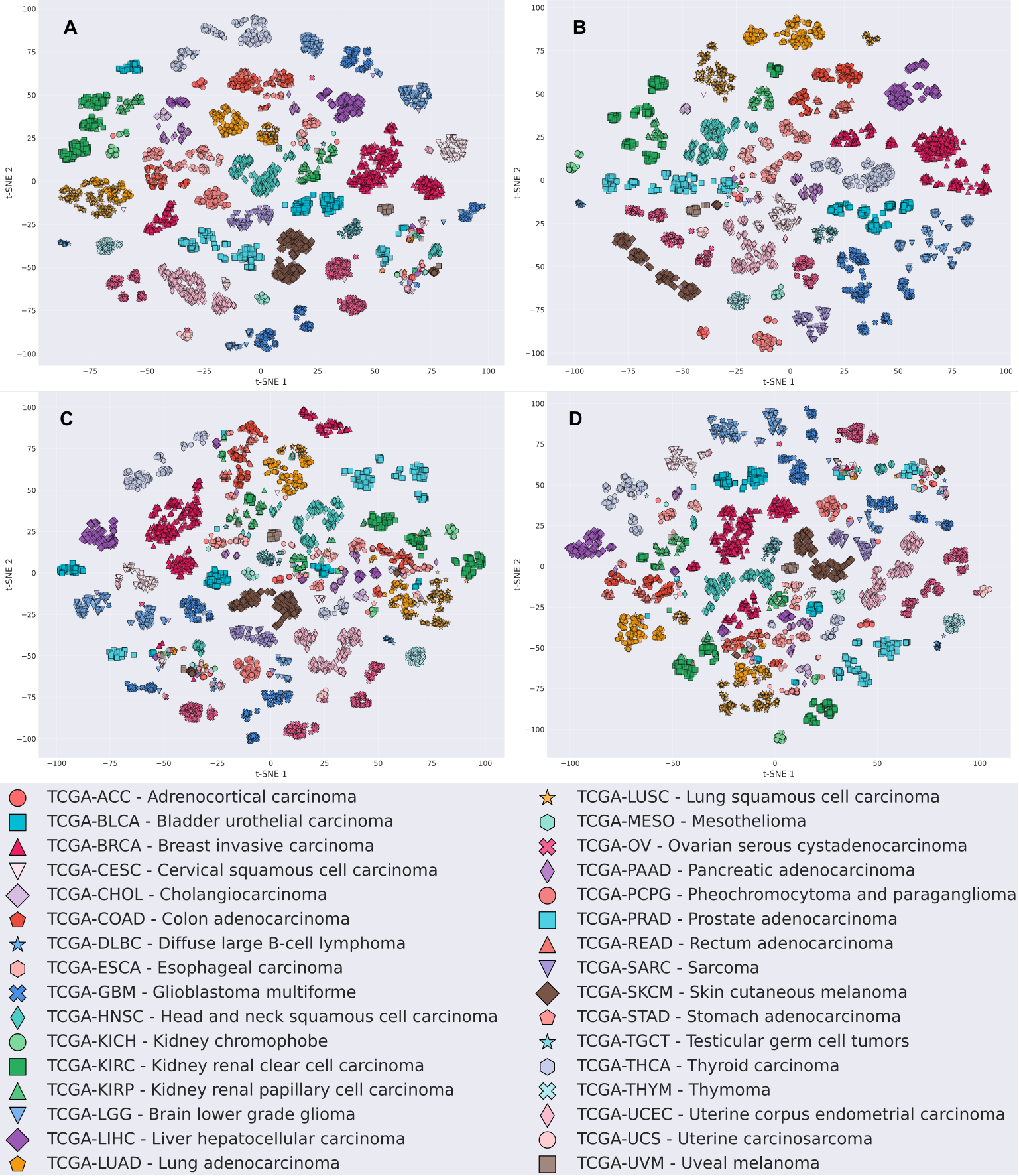}
    \caption{t-SNE visualization of clinical text embeddings generated by four language models, colored by TCGA cancer type. (A) GatorTron embeddings (AMI: 0.71, NMI: 0.75), (B) Qwen3 embeddings (AMI: 0.79, NMI: 0.82), (C) Med-Gemma embeddings (AMI: 0.59, NMI: 0.65), and (D) Llama-3.2 embeddings (AMI: 0.60, NMI: 0.66). Each point represents a patient, with colors and symbols indicating one of 33 cancer types from TCGA. The visualizations demonstrate varying dgrees of cancer-type clustering across different language models, with Qwen3 achieving the highest clustering quality (AMI: 0.79) followed by GatorTron (AMI: 0.71), while Med-Gemma and Llama-3.2 show moderate clustering performance, reflecting each model's ability to capture cancer-specific information from clinical data.}
    \label{fig:clinical-cancer-type-tsne}
\end{figure}

We conducted a systematic evaluation of four LLMs for generating clinical text embeddings: GatorTron (medical encoder model) \cite{gatortron}, Qwen3 (general-purpose encoder) \cite{qwen}, Med-Gemma (medical decoder model) \cite{medgemma}, and LLaMA (general-purpose decoder) \cite{llama}. These models were evaluated using clinical notes and pathology reports from the TCGA dataset across three key tasks: cancer type classification, patient retrieval, and overall survival prediction. Fig.~\ref{fig:clinical-cancer-type-tsne} presents t-SNE visualizations of the embedding spaces produced by each model, illustrating distinct differences in how various architectures and training strategies encode cancer-specific information. The degree of cluster separation and organization varied markedly between models, with direct implications for downstream task performance.


Table \ref{tab:classification_results} presents the classification accuracy across all four models for both text modalities. Clinical text embeddings demonstrated better performance, with accuracies exceeding 98\%. Notably, the general-purpose encoder Qwen achieved the highest accuracy (99.95\%), suggesting that broad language understanding capabilities translate effectively to clinical text representation. Cancer type classification using pathology reports proved more challenging, with accuracies ranging from 78.41\% to 84.90\%, where decoder-only models showed a slightly better performance.

\begin{table}[htbp]
\centering
\caption{Cancer type classification performance (mean accuracy ± std) across 10 runs of Random Forest classifier for 33 TCGA cancer types using embeddings from clinical data and pathology reports.}
\label{tab:classification_results}
\begin{tabular}{@{}lcccc@{}}
\toprule
\textbf{Model} & \textbf{Architecture} & \textbf{Domain} & \textbf{Clinical Data} & \textbf{Pathology Reports} \\
\midrule
GatorTron & Encoder & Medical & 98.66 ± 0.11 & 78.41 ± 0.88 \\
Qwen & Encoder & General & \textbf{99.95 ± 0.04} & \textbf{84.90 ± 0.60} \\
Med-Gemma & Decoder & Medical & 99.52 ± 0.09 & 84.05 ± 0.72 \\
Llama-3.2 & Decoder & General & 99.49 ± 0.16 & 82.45 ± 0.61 \\
\bottomrule
\end{tabular}
\end{table}


We evaluated the quality of embeddings through retrieval tasks, measuring precision@k for identifying patients with the same cancer type. As shown in Table \ref{tab:llm_retrieval_results}, clinical embeddings demonstrated superior retrieval performance, with Qwen achieving near-perfect precision@10 (99.57\%). The substantial performance gap between clinical and pathology report embeddings highlights the structured nature of clinical narratives compared to the more heterogeneous pathology reports.

\begin{table}[htbp]
\centering
\caption{Retrieval performance metrics for patient similarity search across 33 TCGA cancer types using embeddings from four language models applied to clinical data and pathology reports. Metrics include precision at k (P@k) for k={1, 10, 50} and adjusted mutual information (AMI).}
\label{tab:llm_retrieval_results}
\begin{tabular}{@{}lcccccccc@{}}
\toprule
& \multicolumn{4}{c}{\textbf{Clinical Data}} & \multicolumn{4}{c}{\textbf{Pathology Reports}} \\
\cmidrule(lr){2-5} \cmidrule(lr){6-9}
\textbf{Model} & \textbf{P@1} & \textbf{P@10} & \textbf{P@50} & \textbf{AMI} & \textbf{P@1} & \textbf{P@10} & \textbf{P@50} & \textbf{AMI} \\
\midrule
GatorTron & 0.988 & 0.964 & 0.902 & 0.970 & 0.814 & 0.703 & 0.564 & 0.748 \\
Qwen & \textbf{0.999} & \textbf{0.996} & \textbf{0.974} & \textbf{0.996} & 0.861 & 0.780 & 0.651 & 0.816 \\
Med-Gemma & 0.989 & 0.955 & 0.861 & 0.955 & 0.830 & 0.713 & 0.529 & 0.758 \\
Llama-3.2 & 0.993 & 0.965 & 0.884 & 0.963 & 0.847 & 0.740 & 0.564 & 0.774 \\
\bottomrule
\end{tabular}
\end{table}


To assess the clinical utility of embeddings, we conducted stratified survival analysis using Cox proportional hazards, random survival forests, and DeepSurv models. Table \ref{tab:llm_survival_results} summarizes the average concordance indices across 33 cancer types. Clinical embeddings consistently achieved C-indices above 0.84, demonstrating strong prognostic value. In contrast, pathology report embeddings showed limited predictive power (C-index ~0.58), suggesting that current language models may not fully capture the prognostic information embedded in pathological descriptions.

\begin{table}[htbp]
\centering
\caption{Average concordance indices for overall survival prediction across 33 TCGA cancer types using embeddings from four language models applied to clinical data and pathology reports. Results show performance for three survival models: Cox PH, RSF, and DeepSurv.}
\label{tab:llm_survival_results}
\begin{tabular}{@{}lcccccc@{}}
\toprule
& \multicolumn{3}{c}{\textbf{Clinical Data}} & \multicolumn{3}{c}{\textbf{Pathology Reports}} \\
\cmidrule(lr){2-4} \cmidrule(lr){5-7}
\textbf{Model} & \textbf{Cox PH} & \textbf{RSF} & \textbf{DeepSurv} & \textbf{Cox PH} & \textbf{RSF} & \textbf{DeepSurv} \\
\midrule
GatorTron & 0.855±0.038 & 0.842±0.040 & \textbf{0.866±0.029} & 0.580±0.071 & \textbf{0.581±0.079} & 0.586±0.073 \\
Qwen & 0.842±0.036 & \textbf{0.850±0.028} & 0.862±0.025 & 0.584±0.080 & 0.570±0.082 & \textbf{0.592±0.068} \\
Med-Gemma & 0.856±0.033 & 0.850±0.035 & 0.862±0.027 & \textbf{0.596±0.077} & 0.566±0.076 & 0.584±0.075 \\
Llama-3.2 & \textbf{0.858±0.034} & 0.842±0.030 & 0.861±0.025 & 0.573±0.066 & 0.572±0.080 & 0.573±0.073 \\
\bottomrule
\end{tabular}
\end{table}


Given the lower baseline performance on pathology reports, we investigated whether task-specific fine-tuning could improve embedding quality. We trained lightweight neural network classifiers on top of frozen embeddings and compared their performance to the baseline random forest approach. As shown in Table \ref{tab:finetuning_results}, fine-tuning yielded substantial improvements across all models, with accuracies increasing by 7.8-12.7 percentage points.

\begin{table}[htbp]
\centering
\caption{Cancer type classification accuracy improvement for 33 TCGA cancer types using pathology report embeddings. Baseline results show Random Forest classifier performance, while fine-tuned results use a neural network classifier trained on frozen embeddings from each language model.}
\label{tab:finetuning_results}
\begin{tabular}{lccc}
\hline
\textbf{Model} & \textbf{Baseline (\%)} & \textbf{Fine-tuned (\%)} & \textbf{Improvement} \\
\hline
GatorTron & 78.41 & 91.07 & +12.66 \\
Qwen & 84.90 & 92.73 & +7.82 \\
Med-Gemma & 84.05 & 94.29 & +10.24 \\
Llama-3.2 & 82.45 & 92.96 & +10.50 \\
\hline
\end{tabular}
\end{table}

\begin{figure}[H]
      \centering
      \includegraphics[width=\textwidth]{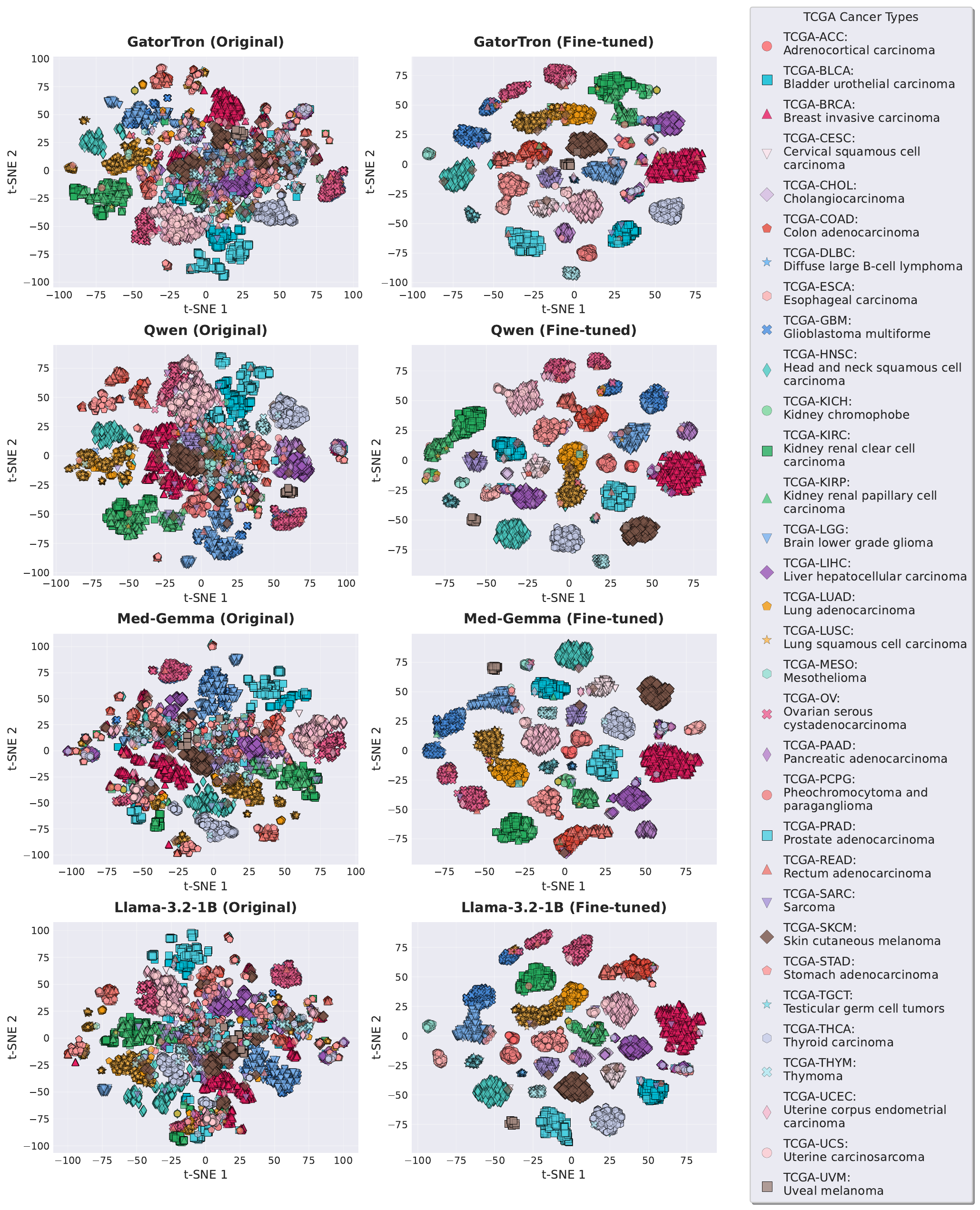}
      \caption{t-SNE visualization of embeddings generated using four language models for the pathology reports. Each point represents a patient colored by TCGA cancer type. Fine-tuned models show improved cluster separation compared to pre-trained models, with AMI scores increasing for all models: GatorTron: 0.35 to 0.91, Qwen3: 0.39 to 0.93, Med-Gemma: 0.26 to 0.93, Llama: 0.32 to 0.93. These results demonstrate the superior cancer type discrimination after LLM fine-tuning.}
      \label{fig:tsne_all_models}
  \end{figure}

Fig. \ref{fig:tsne_all_models} shows a visual representation of the embedding space using t-SNE projections before and after fine-tuning the LLMs using task-specific data. The fine-tuned embeddings demonstrate markedly improved cluster separation. This improvement may potentially translate to better downstream task performance while maintaining computational efficiency through the parameter-efficient fine-tuning approaches.

\section{Discussion}\label{sec:discussion}

In this study, we introduced HONeYBEE, a modular, open-source framework designed to generate unified, patient-level embeddings from diverse biomedical data modalities, including clinical text, pathology reports, radiology images, WSIs, and molecular profiles. By leveraging domain-specific FMs and supporting flexible multimodal fusion strategies, HONeYBEE enables comprehensive representation learning suitable for a wide range of oncology tasks. Using multimodal data from over 11,400 patients spanning 33 cancer types in TCGA, we demonstrated that HONeYBEE-generated embeddings are robust, generalizable, and consistently effective across classification, OS prediction, and patient retrieval tasks. These findings establish HONeYBEE as a scalable and extensible platform for multimodal oncology research. The framework and resulting embeddings are publicly available.

Beyond its empirical performance, HONeYBEE offers technical advantages through its unified architecture. Existing multimodal AI development often relies on fragmented pipelines requiring manual coordination of modality-specific tools such as OHDSI for structured clinical data, PyDicom for radiology imaging, OpenSlide for pathology images, and TIAToolbox for histopathology processing \cite{ohdsi, pydicom, openslide, tiatoolbox}. This fragmentation introduces significant technical overhead when harmonizing cross-modal metadata and managing inconsistent APIs. HONeYBEE resolves these challenges through a modular, plug-and-play framework that standardizes data ingestion, embedding generation, and fusion via a unified API and encoder architecture \cite{MINDS}. Leveraging pretrained FMs from the Hugging Face ecosystem \cite{boehm2022harnessing}, HONeYBEE supports scalable and efficient embedding generation across diverse modalities, while accommodating missing data without requiring complete-case cohorts.

Our comprehensive evaluation revealed that clinical embeddings consistently dominated across clustering, classification, retrieval, and survival prediction tasks. This finding likely reflects the curated nature of TCGA clinical documentation, where key diagnostic variables, such as cancer subtype, stage, grade, and molecular markers, are systematically extracted by clinical experts and recorded as structured narratives. These expert-curated summaries encapsulate prognostically relevant information that remains distributed across raw imaging, pathology, and molecular data, explaining why clinical embeddings achieved the strongest cancer-type separability (NMI: 0.7448, AMI: 0.702), highest classification accuracy (98.5\%), and top retrieval precision@10 (96.4\%). Clinical embeddings also consistently achieved the highest C-indices in survival prediction across 27 of 33 cancer types. These results underscore the critical role of structured clinical documentation in oncology AI workflows, at least in curated datasets like TCGA.

While clinical embeddings outperformed other modalities in most tasks, multimodal integration demonstrated significant value in specific contexts. Particularly in OS prediction, multimodal fusion using concatenation improved prognostic performance for certain cancer types, such as TCGA-UCS, TCGA-UVM, and TCGA-THYM, where complementary information from molecular, pathology, and imaging modalities contributed to enhanced predictions. 

More broadly, fusion strategies like concatenation consistently outperformed weaker individual modalities (e.g., radiology, WSI) across clustering and classification tasks. These findings suggest that while clinical data dominate in curated datasets, multimodal fusion holds critical importance for developing robust models in real-world settings where clinical documentation may be incomplete, inconsistent, or less structured.

In our comparative evaluation of four language models (GatorTron, Qwen3, Med-Gemma, and LLaMA-3.2), we found that general-purpose models like Qwen3 achieved superior clustering (AMI: 0.79, NMI: 0.82), classification accuracy (99.95\%), and retrieval precision compared to domain-specific models like GatorTron and Med-Gemma when applied to clinical text. This suggests that large-scale general-purpose models may encode broader semantic structures that transfer effectively to clinical tasks, especially when applied without further fine-tuning. However, domain-specific models like GatorTron still demonstrated strong and consistent performance, particularly in OS prediction tasks, reinforcing the relevance of clinical pretraining for capturing prognostic signals. These findings indicate that while general-purpose models can outperform specialized architectures in certain tasks, optimal model selection depends on task-specific objectives and data characteristics.

Additionally, fine-tuning significantly enhanced the performance of language models on pathology reports, a modality where baseline embeddings performed suboptimally. Across all models, fine-tuned embeddings improved cancer type classification accuracy by 7.8–12.7 percentage points and increased AMI scores from approximately 0.35–0.39 to 0.91–0.93. This substantial gain underscores the importance of task-specific adaptation when processing heterogeneous and less-structured modalities like pathology narratives. These results highlight that parameter-efficient fine-tuning can effectively bridge the performance gap between raw and curated text, suggesting that real-world deployment of clinical AI systems should incorporate fine-tuning pipelines to optimize embeddings for downstream oncology tasks.

In summary, HONeYBEE establishes a versatile, open-source foundation for multimodal representation learning in oncology, validated across diverse tasks and data modalities. Our findings emphasize the dominant role of clinical text in structured datasets like TCGA, while highlighting the potential of multimodal fusion and targeted fine-tuning to improve performance in real-world, less-curated environments. Future work will focus on extending HONeYBEE to prospective clinical data, evaluating embedding utility for treatment response prediction, and integrating additional modalities such as genomics, radiomics, and wearable sensor data. By facilitating modular, task-agnostic embedding generation, HONeYBEE offers a scalable platform for advancing precision oncology research and clinical applications.

Furthermore, HONeYBEE’s unified patient-level embeddings enable practical applications across oncology research and clinical workflows. These include biomarker discovery, cross-modal hypothesis generation, patient similarity search, cohort identification, and clinical decision support \cite{boehm2022harnessing, azuaje2019artificial, ovchinnikova2024overcoming, truhn2024large, fang2021deepan}. Generated embeddings are compact and interoperable with vector database systems such as FAISS and Annoy \cite{faiss, annoy}, supporting deployment in retrieval-augmented generation, risk stratification, and clinical triage pipelines. Critically, since embeddings can be generated locally without sharing raw data, HONeYBEE facilitates federated learning collaborations \cite{Kout-fm-review}, promoting secure, multi-institutional model development across diverse oncology datasets.

\section{Methods}\label{sec:methods}
HONeYBEE is a modular, open-source framework designed to enable scalable, reproducible, and standardized preprocessing of multimodal oncology data for downstream AI application development, machine learning, and representation learning tasks. The framework provides dedicated processing pipelines for four primary biomedical data modalities: whole slide histopathology images, radiology imaging, molecular profiles, and clinical text. Each modality-specific pipeline incorporates state-of-the-art preprocessing techniques tailored to the data type and is designed to generate high-quality, fixed-length data representations (embeddings) using domain-specific FMs. These embeddings are stored in a structured and accessible format to support classification, retrieval, survival analysis, and other downstream applications.

The HONeYBEE architecture supports the integration of both public and institutional datasets and emphasizes interoperability through standardized APIs and metadata handling. Pretrained and custom models are supported via Hugging Face, with any open-source FM whose weights are available on Hugging Face being easily replaceable with the existing one seamlessly for the given modality. Outputs are organized for compatibility with modern AI pipelines. Fig. \ref{fig:overview} provides an overview of the framework’s architecture, highlighting the sequence of data ingestion, preprocessing, embedding generation, and downstream application.

\subsection{Pathology Image Processing}\label{subsec:wsi}
\begin{figure}[htpb]
    \centering
    \includegraphics[width=\textwidth]{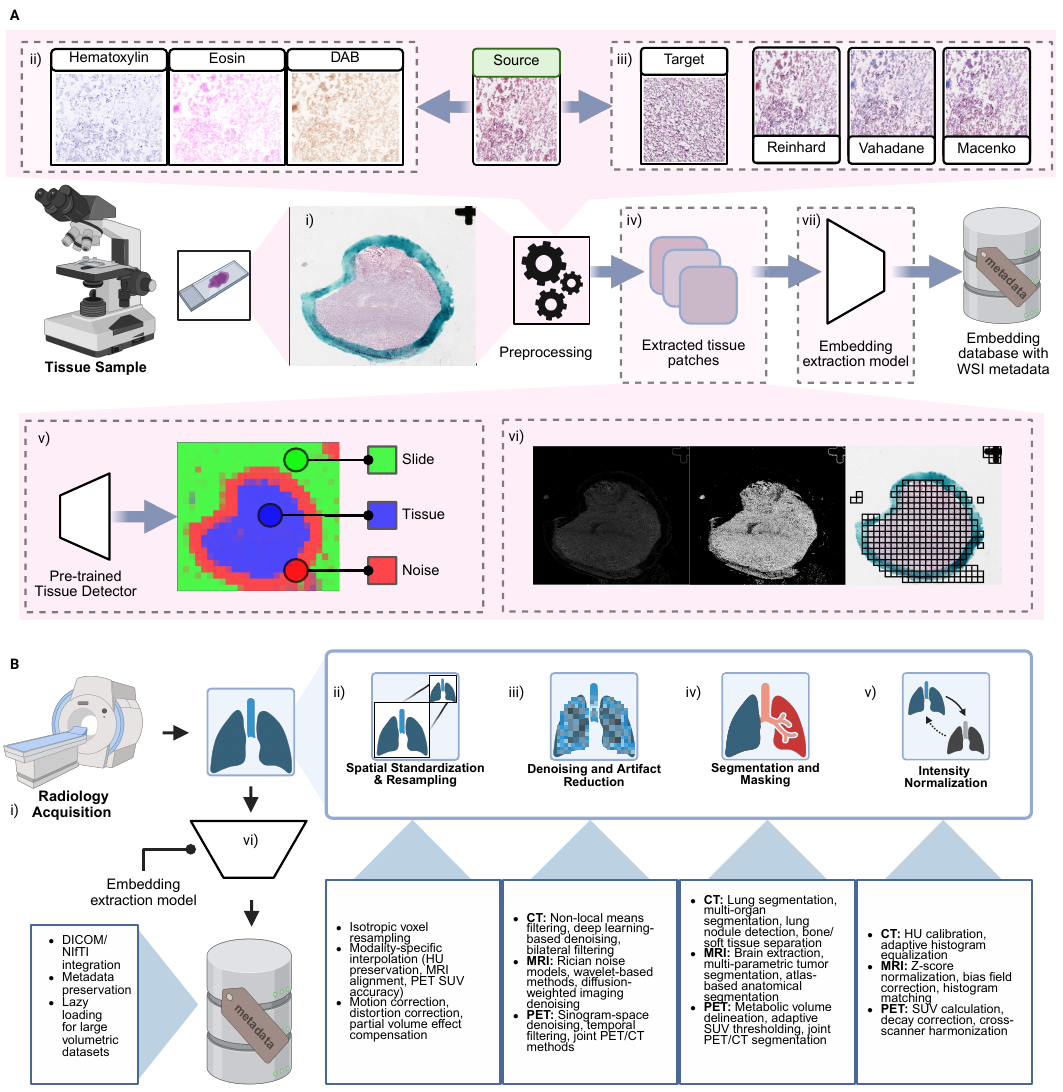}
    \caption{
    \textbf{Pathology and radiology imaging data processing pipelines implemented in HONeYBEE}. (A) Whole Slide Image Processing. (i) Tissue samples are digitized into gigapixel WSIs using slide scanners. (ii) Stain normalization is applied to reduce inter-slide variability caused by differences in hematoxylin, eosin, or DAB staining. Methods include Reinhard, Vahadane, and Macenko normalization techniques. (iii) The source WSI undergoes preprocessing, including tissue segmentation and filtering. (iv) Tissue regions are divided into smaller, information-rich patches for analysis. (v) A pretrained tissue detector model classifies regions as slide background, tissue, or noise to identify high-quality tissue areas. (vi) Grid-based patch extraction is performed over valid tissue regions. (vii) A pretrained embedding extraction model processes the tissue patches to generate fixed-length vectors. These embeddings, along with metadata, are stored in a structured database for downstream AI applications. \textbf{(B) Radiological Image Processing.} (i) Radiology images are acquired and ingested in DICOM or NIfTI format. (ii) Spatial standardization and resampling are applied to harmonize voxel spacing and orientation. (iii) Denoising and artifact reduction methods, such as non-local means or deep learning-based techniques, improve signal quality. (iv) Segmentation models isolate anatomical structures or lesions. (v) Intensity normalization ensures consistency across patients and scanners. (vi) Preprocessed images are passed through an embedding extraction model, and the resulting feature vectors are stored with metadata. This standardized pipeline enables downstream analysis such as classification, retrieval, and prognosis modeling across both pathology and radiology imaging modalities.
    }
    \label{fig:wsi-rad}
\end{figure}

Whole Slide Images (WSIs) present unique computational challenges due to their extreme size, multi-resolution pyramid structure, and vendor-specific file formats. HONeYBEE addresses these challenges through an integrated processing pipeline specifically designed for computational pathology workflows. Our implementation leverages GPU acceleration to efficiently handle WSI loading, preprocessing, tissue detection, normalization, and feature extraction, as illustrated in Fig. \ref{fig:wsi-rad}A.

HONeYBEE utilizes CuImage \cite{cucim} for efficient loading and handling of WSIs, supporting multiple vendor-specific formats, including Aperio SVS, Philips TIFF, and generic tiled multi-resolution RGB TIFF files. CuImage provides GPU-accelerated I/O capabilities through CUDA \cite{cucim}, enabling fast loading of specific regions of interest while managing memory constraints inherent in working with gigapixel-scale images. The library supports various compression schemes, including JPEG, JPEG2000, LZW, and Deflate, providing flexibility in handling different WSI formats while maintaining processing speed.

The framework implements metadata handling to preserve critical image properties and structure. This includes maintaining physical spacing information in micrometers, coordinate system specifications, and vendor-specific metadata such as objective magnification and microns-per-pixel ratios. For formats like Aperio SVS, detailed scan parameters, and ICC color profiles are preserved, allowing for accurate spatial measurements and color reproduction in downstream analysis tasks.

HONeYBEE's multi-resolution pyramid management allows efficient access to WSI data at different magnification levels. The system maintains resolution information, including dimensions, downsampling factors, and tile specifications for each level. This enables memory-efficient implementation of various analysis strategies, from rapid whole-slide thumbnail generation to high-resolution region analysis. The framework implements on-demand loading of specific regions at desired resolution levels, supporting both CPU and GPU memory targets through device specification. When available, the system can leverage NVIDIA GPUDirect Storage for direct data transfer from storage to GPU memory, further reducing I/O bottlenecks \cite{gpudirect_storage}.

Memory efficiency is achieved through intelligent region reading strategies that maintain coordinate consistency across resolution levels. Users can specify locations in base-resolution coordinates while accessing data at any pyramid level, simplifying the development of multi-scale analysis pipelines. The implementation includes automatic memory management through caching strategies, optimizing performance while preventing memory overflow when working with multiple large WSIs simultaneously.

Effective tissue detection and patch extraction are essential preprocessing steps for computational pathology analysis. As shown in Fig. \ref{fig:wsi-rad}, HONeYBEE implements two distinct approaches for tissue detection: a classical threshold-based method using Otsu's algorithm \cite{otsu1975threshold} and a deep learning-based approach utilizing a pretrained DenseNet Slidl model \cite{berman2023slidl}.

The Otsu-based method operates on the assumption that tissue regions exhibit distinct intensity characteristics compared to the background. After converting the image to grayscale, the algorithm automatically determines an optimal threshold that maximizes the variance between tissue and background classes. As shown in Fig. \ref{fig:wsi-rad}, this approach generates a gradient magnitude map and binary tissue mask, followed by grid-based patch extraction. While computationally efficient, this method works best with images that have good contrast between tissue and background regions.

The deep learning-based approach employs a pretrained DenseNet model that classifies image patches into three categories: tissue, background, and noise. This model, originally trained by Slidl \cite{berman2023slidl}, provides more robust tissue detection and can identify artifacts such as pen markings that should be excluded from analysis. Fig. \ref{fig:wsi-rad} demonstrates the model's capability to distinguish between actual tissue regions, background, and artifacts like pen markings, enabling more precise patch extraction. However, even this sophisticated approach can face challenges with certain types of slides. 

Following tissue detection, HONeYBEE implements an intelligent patch extraction strategy that maximizes the coverage of relevant tissue regions while minimizing computational overhead. The framework divides detected tissue regions into standardized patches with configurable size parameters (typically 256×256 or 512×512 pixels). Patches are filtered based on tissue content thresholds to ensure meaningful analysis, and their coordinates are maintained relative to the original WSI for spatial reference. The extracted patches can be efficiently loaded into memory as needed, enabling scalable processing of large whole slide images while maintaining spatial context for downstream analysis tasks.

For whole-slide images, patch-level embeddings from the UNI model are aggregated to patient-level representations using mean pooling across all tissue-containing patches. While this approach may not capture the full complexity of histopathological patterns compared to specialized Multiple Instance Learning methods, it provides a standardized baseline representation compatible with HONeYBEE's unified embedding framework.

Histopathological image appearance varies significantly due to differences in tissue preparation, scanner specifications, and laboratory protocols. These variations negatively impact both visual interpretation and computational analysis. HONeYBEE addresses this challenge by implementing three state-of-the-art stain normalization methods, as illustrated in Fig. \ref{fig:wsi-rad}B:

\begin{enumerate}
    \item Reinhard Normalization: This technique performs color normalization by matching statistical properties of the source image to a target image in LAB color space. While computationally efficient, it does not explicitly account for specific hematoxylin and eosin stain characteristics \cite{reinhard}.
    \item Macenko Normalization: This method estimates stain vectors through singular value decomposition in optical density space, enabling separation of hematoxylin and eosin contributions before normalization. This approach better preserves relative staining patterns while standardizing overall appearance \cite{macenko}.
    \item Vahadane Normalization: Using sparse non-negative matrix factorization, this technique decomposes images into stain density maps, offering robust stain separation even with varying tissue characteristics or additional stain presence \cite{vahadane}.
\end{enumerate}

All three normalization methods in HONeYBEE are GPU-accelerated, providing significant performance improvements over CPU-based alternatives (average 8-12× speedup on typical hardware configurations). Our implementation maintains consistency in normalization results across different image scales through automated parameter adjustment and handles memory management for large WSIs through efficient tiling strategies. A comprehensive evaluation of how these normalization methods impact the performance of different foundation models is presented in the Supplementary Materials (see Supplementary Note - Staining impact of performance, Supplementary Fig. 10 and 11, and Supplementary Table 7 and 8), where we demonstrate that the benefit of stain normalization varies significantly based on model architecture and training strategy.

Beyond normalization, quantitative tissue analysis often requires the isolation of individual stain components. HONeYBEE implements color deconvolution methods to decompose H\&E stained images into constituent components based on the Beer-Lambert law of light absorption, as shown in Fig. \ref{fig:wsi-rad}A.

Our implementation converts RGB images to Hematoxylin-Eosin-DAB (HED) color space through calibrated deconvolution matrices that account for the characteristic absorption spectra of each stain. This transformation enables separate analysis of (i) the hematoxylin component, Highlighting nuclear structures; (ii) the eosin component, Emphasizing cytoplasmic and stromal features; and (iii) the DAB component, Isolating immunohistochemical signals when present.

The framework provides bidirectional conversion between RGB and HED color spaces, allowing researchers to analyze separated stain channels and reconstruct normalized images with modified stain characteristics. These operations are GPU-accelerated and process large image regions through automated tiling strategies to maintain memory efficiency.

\subsection{Radiology Imaging Processing}\label{sec:rad_proc}
Medical imaging plays a central role in oncology for diagnosis, staging, treatment planning, and response assessment. HONeYBEE implements comprehensive processing pipelines for major radiological modalities, including Computed Tomography (CT), Magnetic Resonance Imaging (MRI), and Positron Emission Tomography (PET). Each modality presents unique characteristics that require specialized preprocessing techniques while maintaining standardized outputs for downstream multimodal integration.

HONeYBEE provides support for standard medical imaging formats through integration with industry-standard Digital Imaging and Communications in Medicine (DICOM) protocols and specialized neuroimaging formats such as Neuroimaging Informatics Technology Initiative (NIfTI) \cite{elhadad2024reduction, levitas2024ezbids}. Our implementation preserves metadata that influences image interpretation and quantitative analysis. For CT imaging, this includes Hounsfield unit (HU) calibration parameters, exposure settings (kVp, mAs, dose), slice thickness, reconstruction kernel information, and scanner-specific calibration factors. MRI sequence metadata preservation encompasses pulse sequence parameters (TR/TE values), field strength, coil configuration, contrast administration details, and acquisition plane and timing information. PET studies metadata retention includes radiopharmaceutical type and injection parameters, standardized uptake value (SUV) conversion factors, attenuation correction methods, and decay correction timing information \cite{nanni2023thirty}.

The framework implements memory-efficient loading strategies through lazy evaluation, allowing researchers to work with large volumetric datasets without excessive memory requirements \cite{schafer2024overcoming}. This is particularly important for multi-sequence MRI studies or time-series imaging data that can exceed several gigabytes per patient. The implementation supports both single-file and multi-file DICOM series, automatically handling series organization, temporal sequencing, and multi-frame images \cite{wulms2023package, dhapola2022scarf}.

Isolating relevant anatomical structures and regions of interest forms a critical preprocessing step for targeted analysis of oncological imaging. HONeYBEE incorporates both classical algorithms and deep learning-based approaches for automated segmentation across different imaging modalities \cite{ma2024segment, jin2023novel}. For CT segmentation, the framework offers automated lung segmentation using threshold-based approaches with morphological refinement, multi-organ segmentation through pre-trained U-Net architectures optimized for thoracic, abdominal, and pelvic regions, automated lung nodule detection using 3D convolutional networks with sensitivity exceeding 85\% for nodules larger than 4mm, and bone and soft tissue separation using HU-based thresholding with anatomical constraints \cite{primakov2022automated}.

MRI segmentation capabilities include brain extraction tools optimized for different MRI sequences (T1, T2, FLAIR), multi-parametric tumor segmentation integrating information across multiple sequences, specialized sequence-specific segmentation models for contrast-enhanced regions, and atlas-based segmentation for standardized anatomical referencing \cite{wasserthal2025researchers}. For PET-based segmentation, the framework implements metabolic volume delineation using fixed and adaptive SUV thresholding, gradient-based segmentation for heterogeneous uptake regions, joint PET/CT segmentation leveraging complementary information from both modalities, and automated detection of hypermetabolic regions with statistical deviation analysis \cite{jin2023novel}.

All segmentation methods are implemented with GPU acceleration when applicable, providing significant performance improvements over CPU-based alternatives \cite{jin2024gpu, rosen2021learning}. The framework maintains original image coordinates and transformation matrices, ensuring accurate spatial registration between original images and segmentation masks.

Image noise and artifacts can significantly impact both visual interpretation and computational analysis of medical images. HONeYBEE implements multiple denoising strategies optimized for specific imaging modalities and noise characteristics. CT denoising approaches include non-local means filtering with optimized parameters for different dose levels, deep learning-based denoising through convolutional neural networks trained on paired low-dose/high-dose CT images, edge-preserving bilateral filtering with automatic parameter tuning, and structural fidelity metrics to ensure critical diagnostic features remain uncompromised \cite{qu2024research}.

MRI denoising implementations feature Rician noise models specifically designed for magnitude MRI data, non-local means filtering adapted for different sequence types, wavelet-based denoising with soft thresholding, and specialized approaches for diffusion-weighted imaging addressing unique noise characteristics \cite{koonjoo2021boosting}. PET denoising capabilities encompass sinogram-space denoising for raw PET data when available, image-space denoising with SUV preservation guarantees, temporal filtering for dynamic PET acquisitions, and joint PET/CT denoising leveraging structural information from CT \cite{lu2023imc}.

Our quantitative evaluations demonstrate that these denoising methods achieve an average 40-60\% reduction in noise levels while maintaining structural similarity indices (SSIM) above 0.92 compared to reference images. Performance metrics for each method are continuously benchmarked across public datasets to ensure optimal parameter selection \cite{sanaei2023employing}.

Variability in spatial resolution and slice thickness across different scanners and protocols necessitates standardization for consistent analysis \cite{ma2024segment}\cite{jin2024gpu}. HONeYBEE implements comprehensive resampling capabilities with modality-specific optimization. Resolution standardization includes isotropic voxel resampling with configurable target dimensions, support for both downsampling and upsampling with appropriate filtering, preservation of spatial relationships and anatomical proportions, and multiple interpolation methods including nearest neighbor, linear, B-spline, and Lanczos \cite{schafer2024overcoming, lee2023comparative}.

CT-specific resampling techniques focus on Hounsfield unit preservation during interpolation, thin-slice to thick-slice conversion with appropriate averaging, axial to coronal/sagittal reformatting with consistent spatial referencing, and partial volume effect compensation for quantitative analysis \cite{mottola2021reproducibility}. MRI-specific resampling addresses multi-sequence alignment to establish spatial correspondence, motion correction between sequences acquired at different timepoints, distortion correction for echo-planar imaging sequences, and field inhomogeneity compensation for accurate spatial registration \cite{bernatz2021impact}. PET-specific resampling includes SUV-preserving interpolation methods, PET/CT alignment with motion compensation, resolution recovery for quantitative accuracy, and partial volume effect correction for small lesions \cite{leijenaar2015effect}.

Our implementation tracks and propagates spatial transformation matrices through all processing steps, enabling accurate coordinate mapping between original and processed images. This spatial consistency is critical for multimodal integration and region-of-interest analysis \cite{marfisi2022image}.

Intensity standardization addresses variability in signal intensities across different scanners, protocols, and patient characteristics. HONeYBEE implements modality-specific normalization techniques that preserve quantitative relationships while enabling consistent analysis. CT intensity normalization includes Hounsfield unit verification and calibration, predefined window/level settings for different anatomical regions (lung, soft tissue, bone), adaptive histogram equalization for enhanced contrast in selected regions, and intensity outlier handling for metal artifacts and beam hardening \cite{ma2024segment, wuschner2021radiation}.

MRI intensity normalization encompasses Z-score normalization on a per-sequence basis, histogram matching to standardized references, bias field correction for signal inhomogeneity, tissue-specific intensity normalization using segmentation priors, and cross-sequence intensity harmonization for multi-parametric analysis \cite{carre2020standardization, daisaki2021usefulness}. PET intensity standardization features SUV calculation with support for different normalization factors (body weight, lean body mass, body surface area), time-decay correction for quantitative accuracy, scanner-specific calibration factor application, and cross-scanner harmonization for multi-center studies \cite{landau2024positron}.

The framework maintains both raw and normalized versions of the imaging data, allowing researchers to access original quantitative values when needed while providing standardized inputs for machine learning algorithms and AI model development \cite{yayon2018intensify3d}.

\subsection{Molecular Data Processing}\label{subsec:molecular_processing}

\begin{figure}[htpb]
    \centering
    \includegraphics[width=0.98\textwidth]{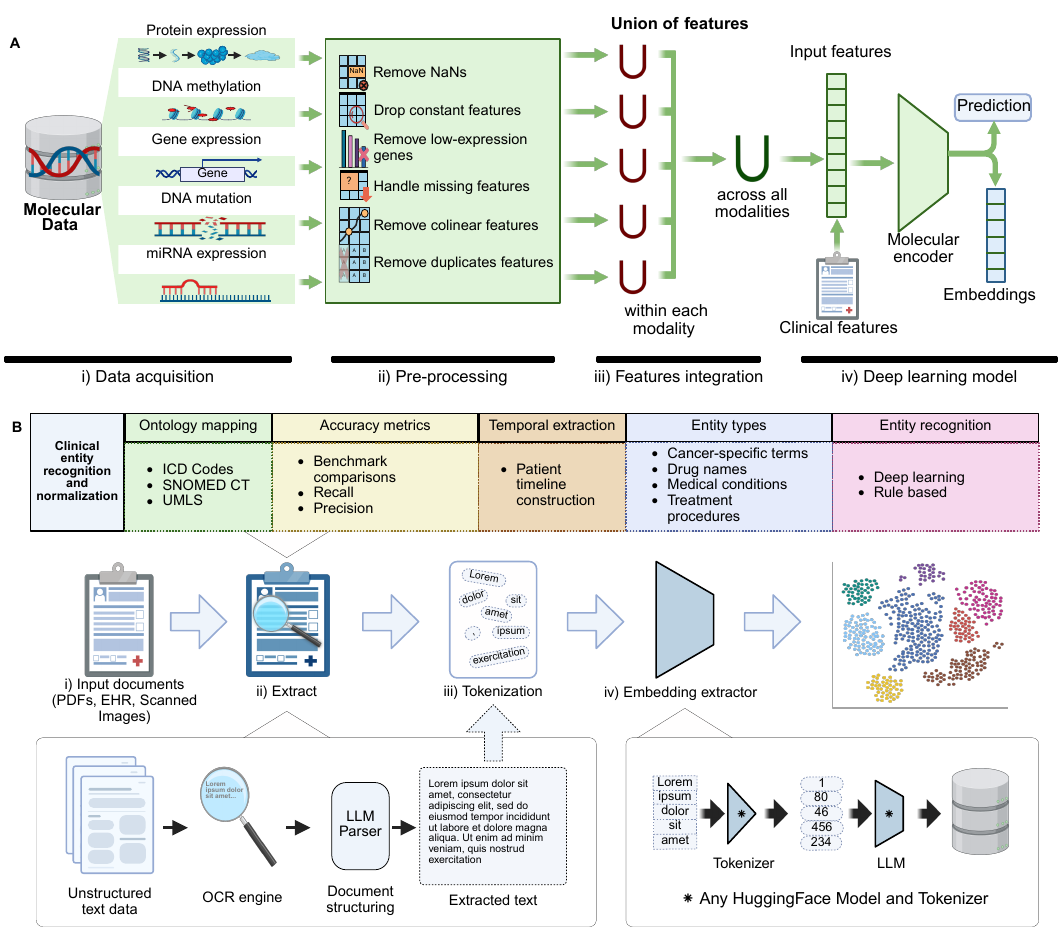}
    \caption{
    \textbf{Molecular and clinical data processing pipelines implemented in HONeYBEE.} (A) Molecular Processing: Multimodal molecular data—such as protein expression, DNA methylation, gene expression, DNA mutation, and miRNA expression—is acquired from public repositories. Preprocessing includes the removal of missing values, constant or duplicate features, low-expression genes, and collinear features. Features are first unified within each modality, then across modalities, and subsequently integrated with clinical data. The combined feature set is passed through a molecular encoder \cite{SeNMo} to generate embeddings or predictions for downstream tasks. \textbf{(B) Clinical Processing:} Structured and unstructured clinical data, including EHRs, PDFs, and scanned reports, undergo entity recognition, normalization, and embedding. Key stages include: (i) document input, (ii) entity extraction and OCR for scanned text, (iii) tokenization using domain-specific tokenizers, and (iv) embedding generation using Hugging Face-compatible language models. The pipeline supports concept mapping (e.g., ICD, SNOMED CT), accuracy benchmarking, timeline construction, and identification of cancer-specific terms, enabling structured integration of clinical narratives for downstream AI applications.}
    \label{fig:cli-mol}
\end{figure}

Molecular data refers to biological information derived from molecules within cells, including DNA, RNA, proteins, and metabolites. Fig.~\ref{fig:cli-mol}A illustrates the molecular data processing pipeline implemented in HONeYBEE, which handles multiple data modalities including DNA methylation, gene expression, protein expression, DNA mutation, and miRNA expression data. Molecular data is widely used in biomedical research, personalized medicine, and multi-omics analyses to understand disease mechanisms, predict outcomes, and develop targeted therapies. It provides insights into cellular functions, genetic variations, and biochemical processes. Common types of molecular data include genomic data (DNA sequences, mutations, variations), transcriptomic data (RNA expressions), proteomic data (protein expressions, modifications, interactions), epigenomic data (DNA methylation, histone modifications), and metabolomic data. Although rich in information, using molecular data for analysis presents several challenges across multiple stages, from data acquisition to interpretation. These challenges include varying complexity across modalities, `\textit{big P-small N}' (very high dimensionality features and low sample size), uncommon standardization, integration, missingness, experimental noise, and high computational and storage demands \cite{liao2007logistic}. Addressing these challenges requires robust statistical methods, computational advancements, standardized protocols, and interdisciplinary collaboration. HONeYBEE framework incorporates all the pre-processing steps as described in the original manuscript of SeNMo framework \cite{SeNMo}, to ensure the integrity of the data while minimizing the complexity in handling the heterogeneous molecular data modalities. As described in Section 2.1.3 of SeNMo \cite{SeNMo}, the molecular data processing pipeline involves five data modalities including DNA methylation, gene expression, protein expression, DNA mutation, and miRNA expression data, acquired from publicly available repositories such as TCGA and University of California, Santa Cruz (UCSC) Xena portal. This data underwent a series of preprocessing steps, including normalization, scaling, dropping constant/ duplicate/ colinear features, removing low-expression genes, removing NaNs, and handling missing data modalities. This pipeline ensures data harmonization, dimensionality reduction, and the generation of contextually enriched embeddings for downstream machine learning and AI applications. These preprocessed feature sets are then unified into a multimodal feature matrix to facilitate pan-cancer analyses. The key steps in feature selection involve the union of features across modalities and cancer types, elimination of redundancies to exclude duplicates/ colinears, multi-modal integration into a single matrix, and addition of clinical covariates. Refer to SeNMo for details of these steps \cite{SeNMo}. For patients with multiple molecular profiles, features are aggregated at the patient level before embedding generation.

\subsection{Clinical Data Processing}\label{sec:clinical_proc}

Clinical data in oncology presents unique challenges due to its heterogeneous nature, spanning structured data elements (laboratory values, vital signs, medications) and unstructured narrative text (clinical notes, radiology reports, pathology reports). HONeYBEE implements processing pipelines to handle both data types while maintaining semantic relationships for clinical analysis, as illustrated in Fig.~\ref{fig:cli-mol}B.

The extraction of clinical text from various document formats represents a critical preprocessing step in clinical data analysis. HONeYBEE supports multiple input formats including PDF, scanned images, and electronic health record (EHR) exports. The framework implements a multi-stage processing pipeline with modality-specific optimizations:
For digitized documents, the framework utilizes a multi-stage optical character recognition (OCR) pipeline utilizing Tesseract OCR with specialized medical dictionaries to improve recognition accuracy of clinical terminology \cite{jantscher2023information}. The pipeline includes image preprocessing (deskewing, noise reduction, binarization), layout analysis to preserve document structure, and post-processing with medical terminology verification to identify and correct common OCR errors in clinical contexts \cite{wang2023extracting}.

For structured EHR data in tabular formats, HONeYBEE implements a conversion pipeline that transforms discrete data elements into standardized key-value pair representations. This process preserves the semantic relationships inherent in the original data structure while enabling unified processing alongside unstructured text \cite{zhang2023synthesize}. Laboratory results, vital signs, medication orders, and other structured elements are converted into a consistent text format (e.g., ``blood\_pressure: 120/80 mmHg", ``white\_blood\_cells: 7.2 K/µL") that can be processed by language models while maintaining clinical meaning and temporal relationships \cite{zhang2023synthesize}.

Document structure analysis is performed to identify and preserve the hierarchical organization of clinical reports, maintaining relationships between sections, subsections, and individual data elements. The framework implements specialized handlers for common clinical document types, including operative reports, pathology reports, and consultation notes, each with specific rules for extracting structured information from semi-structured text \cite{jantscher2023information}. The framework implements quality control measures to flag problematic documents and provides confidence scores for extracted text segments, enabling downstream processes to account for potential extraction errors \cite{wang2023extracting}.

Following successful text extraction, HONeYBEE implements tokenization through integration with the Hugging Face Transformers library, enabling compatibility with state-of-the-art clinical language models. The framework supports multiple pretrained tokenizers optimized for biomedical text, including BioClinicalBERT, PubMedBERT, GatorTron, and ClinicalT5, ensuring appropriate handling of domain-specific vocabulary and clinical narratives \cite{yang2022large, hassan2024optimizing}. The tokenization pipeline processes extracted clinical text through several stages:

\begin{enumerate}
    \item Preliminary cleaning to standardize whitespace, handle special characters, and normalize common clinical abbreviations
    \item Text segmentation into appropriate units (sentences, paragraphs, sections) based on document structure
    \item Subword tokenization using model-specific strategies (WordPiece, Byte-Pair Encoding, SentencePiece)
    \item Special token handling for model-specific requirements (e.g., [CLS], [SEP], [MASK])
    \item Sequence length management with configurable strategies for truncation and sliding windows
\end{enumerate}

To accommodate long clinical documents that exceed typical model input token sequence lengths (generally 512-1024 tokens), HONeYBEE implements multiple strategies: (i) sliding window tokenization with configurable overlap percentages (typically 10-20\%), (ii) hierarchical document segmentation preserving section boundaries, (iii) important segment identification using clinical term density heuristics, and (iv) document summarization for extremely long texts using extractive methods.

The framework's tokenization implementation is optimized for batch processing, enabling efficient handling of large clinical document collections while maintaining memory efficiency through dynamic batch sizing based on available computational resources \cite{alba2025foundational}.


Identification and standardization of clinical concepts are essential for structured analysis of medical narratives. HONeYBEE implements a comprehensive clinical entity recognition pipeline utilizing both rule-based and deep learning approaches\cite{wang2021nero, hassan2024optimizing}.

Our implementation includes integration with medical ontologies and terminologies including SNOMED CT, RxNorm, LOINC, and ICD-O-3, enabling normalization of extracted clinical entities to standardized concept identifiers. This normalization process addresses challenges such as synonymy (multiple terms for the same concept), abbreviation expansion based on context, and disambiguation of terms with multiple potential meanings\cite{wang2021nero}.

For cancer-specific entity recognition, the framework implements specialized models for extracting and normalizing\cite{capper2024automated}: (i) tumor characteristics (histology, grade, size, invasiveness), (ii) staging information (TNM parameters, stage groupings), (iii) biomarker status (receptor expression, molecular alterations), (iv) treatment details (surgical procedures, chemotherapy regimens, radiation protocols), and (iv) response assessment (RECIST criteria, clinical response categories).

Temporal information extraction identifies and normalizes dates, durations, and relative time expressions within clinical narratives. This enables the construction of longitudinal patient timelines with precise sequencing of diagnostic, treatment, and follow-up events critical for oncology research\cite{sun2013temporal, stubbs2015creation}.

The extracted and normalized entities maintain provenance links to their source documents and original text spans, facilitating verification and quality assessment of automated extractions. Performance metrics for entity recognition are continuously benchmarked across public datasets\cite{wang2021nero, hassan2024optimizing}.

\subsection{Embedding Generation and Storage}\label{subsec:embeddings}
Each preprocessed data sample is passed through pre-trained FM, which produces a fixed-length embedding vector that varies in shape depending on the model architecture. HONeYBEE utilizes GPU acceleration and distributed computing, when available, to efficiently generate embeddings for large-scale datasets. The generated embeddings, along with associated metadata, are stored in a structured format to facilitate downstream tasks such as similarity search, clustering, and AI and machine learning model training. HONeYBEE employs efficient data compression and indexing techniques to optimize the storage and retrieval of high-dimensional embeddings. As an example, the processed molecular data is passed through the SeNMo model, a self-normalizing deep learning encoder, to generate latent feature vectors (embeddings). These embeddings capture critical biological patterns and are suitable for tasks such as survival analysis, cancer subtype classification, and biomarker discovery.

The generated embeddings and tabular data are stored using the Hugging Face datasets library, which provides a standardized interface for data access and processing \cite{hfdatasets}. The datasets are organized into a structured format, containing the embeddings, metadata, and labels (if available). PyTorch DataLoaders are employed to efficiently load and iterate over the datasets during model training and evaluation, handling tasks such as batching, shuffling, and parallel processing. Additionally, HONeYBEE datasets can be integrated into vector databases such as Faiss and Annoy \cite{faiss, annoy} to enable fast similarity search, nearest neighbor retrieval, and clustering on the high-dimensional embedding vectors. These databases are optimized for efficient querying and retrieval of embeddings based on similarity metrics such as Euclidean distance or cosine similarity \cite{wu2024forb}. By deploying vector databases, researchers can quickly identify similar samples, perform data exploration, and retrieve relevant subsets based on embedding similarity, facilitating various downstream tasks such as retrieval augmented generation (RAG) \cite{yang2024leandojo}. The structured storage and accessibility components of HONeYBEE ensure that the generated embeddings and associated data are readily available for researchers and practitioners to use in their AI and machine learning pipelines and downstream applications. This research utilized publicly available data from TCGA, which has been previously collected with appropriate ethical approvals and patient consent. No additional human subjects were involved in this study, and therefore, no additional ethics approval was required. The TCGA data were obtained through authorized access and used in accordance with the data use agreements.

\backmatter

\bmhead{Supplementary information}
The online version contains supplementary material available at [DOI will be inserted by publisher].


\bmhead{Acknowledgements}
This research was supported by NSF Awards 2234836 and 2234468 and NAIRR pilot funding.

\section*{Declarations}

\subsection*{Data availability}
The datasets generated during the current study are available in the Hugging Face repository, \url{https://huggingface.co/datasets/Lab-Rasool/TCGA}. The raw data used in this study was obtained from The Cancer Genome Atlas (TCGA) through the MINDS framework \cite{MINDS}. A comprehensive collection of the datasets and models used in this study is available at \url{https://huggingface.co/collections/Lab-Rasool/honeybee-66b9f15e8b908c4eeeec9cb4}.

\subsection*{Code availability}
The source code for the HONeYBEE framework is publicly available on GitHub at \url{https://github.com/lab-rasool/HoneyBee} under an open-source license. Additionally, the MINDS codebase used for data extraction is available at \url{https://github.com/lab-rasool/MINDS}. The HONeYBEE paper preprint is available at \url{https://arxiv.org/abs/2405.07460}.

\subsection*{Authors' contributions}
AT and AW contributed equally to this work. AT led the development of the multimodal embedding framework, designed and implemented the neural network architectures, and conducted computational experiments. AW led the molecular processing and organization efforts and contributed to the molecular model development. MBS provided clinical expertise, validated the clinical relevance of the findings, and contributed to the interpretation of results. YY provided guidance on algorithm design and statistical analysis. GR conceived and supervised the study, secured funding, and provided overall direction. All authors contributed to the manuscript writing and revision.

\subsection*{Competing interests}
The authors declare that they have no competing interests.







\newpage
\bibliography{sn-bibliography}

\end{document}